# Genetic Programming for Document Segmentation and Region Classification Using Discipulus

Priyadharshini N
Research Scholar
PSGR Krishnammal College for Women
Coimbatore, India

Vijaya MS
Associate Professor
GR Govindarajulu School Of Applied Computer Technology
Coimbatore, India

*Abstract*— Document segmentation is a method of rending the document into distinct regions. A document is an assortment of information and a standard mode of conveying information to others. Pursuance of data from documents involves ton of human effort, time intense and might severely prohibit the usage of data systems. So, automatic information pursuance from the document has become a big issue. It is been shown that document segmentation will facilitate to beat such problems. This paper proposes a new approach to segment and classify the document regions as text, image, drawings and table. Document image is divided into blocks using Run length smearing rule and features are extracted from every blocks. Discipulus tool has been used to construct the Genetic programming based classifier model and located 97.5% classification accuracy.

## I. INTRODUCTION

Document segmentation is defined as a method of subdividing the document regions into text and non-text regions. A non-text region includes images, drawings, rules etc. Document segmentation plays a significant role in document analysis, because every day, millions of documents including technical reports, government files, Newspapers, books, magazines, letters, bank cheque, etc,. have to be processed. A lot of time, effort and money will be saved if it can be executed automatically. Automation of document analysis involves region extraction, identification of type of region and finally processing of each region separately. Document segmentation does the work of identifying the type of region. The documents contain both text and non-text regions. In order to process each region, the document should be segmented and then fed into the respective system for processing. For example, text regions are processed using OCR system.

It converts text region into machine-readable form and non-text regions are conserved for processing such as compression, enhancement, recognition, and storage etc. Some document segmentation applications are field extraction and recognition, word searching, logo detection, retrieving imaged documents in digital libraries, retrieval of documents containing tables or drawings. Also document segmentation is being adopted in postal industry where the address fields have to be identified before being sent to OCR readers and stamps have to be recognized.

Generally document regions are segmented in two ways namely, geometric and logical based segmentation. In geometric based segmentation, the document is segmented upon its geometric structure such as text and non-text regions. Whereas in logical segmentation the document is segmented upon its logical labels assigned to each region of the document such as title, logo, footnote, caption, etc., [2].

Till now lots of methods have been proposed for document segmentation in the literature. Document segmentation techniques are broadly classified into three categories: top-down, bottom-up and hybrid approach [2]. A top-down approach repetitively segments the document image into smaller regions until further it cannot be segmented. Run-length smearing algorithm, projection profile methods, Fourier transforms etc., [4, 5, 7] are the methods which make use of top down approach. A bottom-up approach begins by merging pixels into characters. Then the characters are merged into words until whole document regions are merged. The methods which follow this approach are connected component analysis [3], run-length smoothing [6], region-growing methods [4], and neural networks [8]. A hybrid approach is the combination of both top down and bottom up approach. Few hybrid based methods are texture based and Gabor filters. The advantage of using top-down approach is, its high speed processing and the drawback is, it cannot process table, improper layout documents and forms.

This research work associates the existing features specified in [4] [6] [8] and proposes few features which subsidizes more in document segmentation. Features such as perimeter/height ratio, energy, entropy are employed. Perimeter/height ratio is defined as a fraction perimeter to the height of the block. A block in document image is a connected component and it is defined as a collection of black runs that are 8-connected. Both perimeter and height of the block diverges in their values. Text blocks have slighter value for perimeter/height ratio when compared to non-text blocks. Thus perimeter/height ratio is essential in classifying the blocks.

Additionally energy and entropy features are used to classify the blocks. Energy and entropy are renowned properties of an image. Energy identifies the uniformity of the image. Whereas, entropy identifies the randomness (texture) of the image. Each block of the document varies in its energy and entropy specifically in case of table, drawings and image blocks. Thus these new features offer a notable influence in document segmentation.

This paper presents a genetic programming based document region classification. As genetic programming progress the classifiers in the manner of a program and measure the final classifier (program) for the classification conclusion this encourages in straightforward and speedy analysis of results. In this paper, genetic programming is implemented using commercial software called Discipulus. Discipulus writes computer programs automatically in most high level languages. Discipulus uses a multi-run evolutionary algorithm to evolve computer programs from the data. These evolved programs are high-precision models built from the data and map the inputs to the output. Unlike statistical techniques like neural networks, decision trees , discipulus builds models without any user tuning. It is self tuning and self parameterizing.

## II. PROPOSED MODEL FOR DOCUMENT SEGMENTATION AND REGION CLASSIFICATION

The proposed approach reduces the computational complexities of supporting procedure after document segmentation such as OCR system for processing text regions and for processing non text region. It reformulates document segmentation problem as classification task and solved using genetic programming. Documents are collected from various journals and features are extracted from document images. The proposed method consists of four phases: preprocessing, segmentation using Run length smearing algorithm (RLSA), labeling connected components and feature extraction. Each phase is described in the following sections and system architecture is shown in Fig. 1.

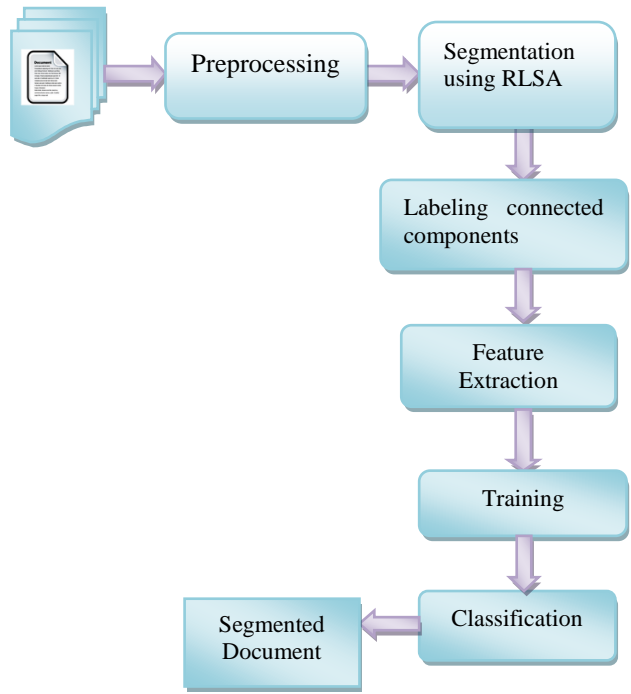

Fig. 1. System Overview

### A. Preprocessing

Preprocessing is a sequence of tasks performed on the document image. It enhances the quality of the image for segmentation. The various tasks performed on the image in preprocessing stage are scanning, binarization, and noise removal.

*1) Scanning:*
The documents are collected from various journals and scanned at 200 dots per inch (dpi).

*2) Binarization:*
It is a process which converts the grayscale image into a binary image using the global threshold method. A binary image has only two values 0 or 1for each pixel. 0 represents white pixel and 1 represents black.

*3) Noise removal:*
From the binarized image, the noise is removed using ostu method. The obtained noiseless image is subjected to RLSA algorithm for further processing.

### B. Segmentation using RLSA algorithm

The RLSA algorithm is used for segmenting the document. The document is subdivided into blocks, where each block contains only one type of data such as text, graphic, halftone image, etc.,.

The Run length algorithm is employed to a binary image where 0 represents white pixels and 1 represents black pixels. The algorithm transforms a binary image **x** into an output image **y** as follows:

1. White pixels in x are replaced with black pixels in y if white runs are less than or equal to a predefined threshold.

2. Black pixels in x are untouched in y.

With a selection of optimal threshold value, the connected areas will form blocks of the same region. The Run Length Algorithm is employed horizontally with horizontal threshold hTh = 300 as well as vertically with vertical threshold vTh = 280 to a document image, producing two individual images. The two images are then combined using logical AND operation. Furthermore, horizontal smearing is applied with horizontal threshold Th = 30 to generate segmented blocks. Fig. 2 shows the segmentation of document image using RLSA algorithm.

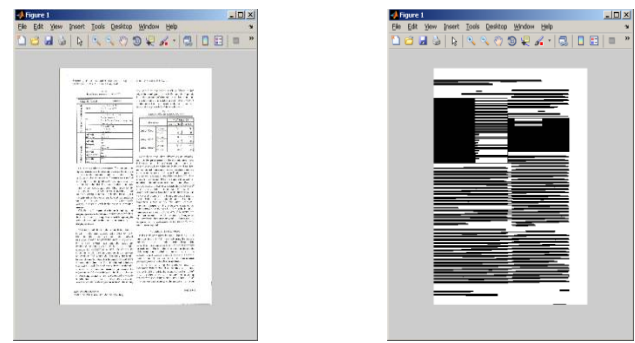

(a)          (b)

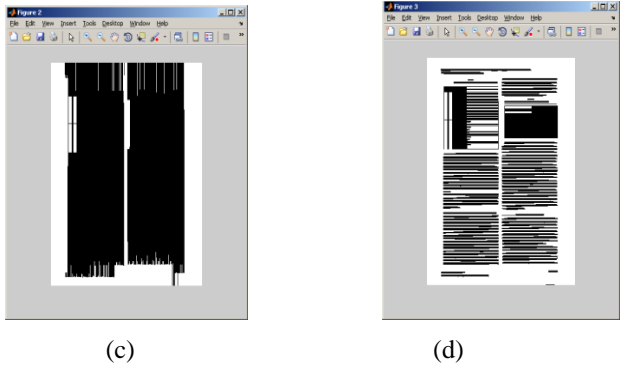

(c)               (d)

Fig. 2. (A) Original Document Image. (B) Result Of Horizontal Smearing. (C) Result Of Vertical Smearing. (D) Final Result Of Segmentation.

### C. Labeling connected components

Labeling is the process of identifying the connected components in an image and assigning each component a unique label an integer number which must be same as connected black runs. Labels are used to differentiate each block. Fig. 3 shows the labeled connected components of document image.

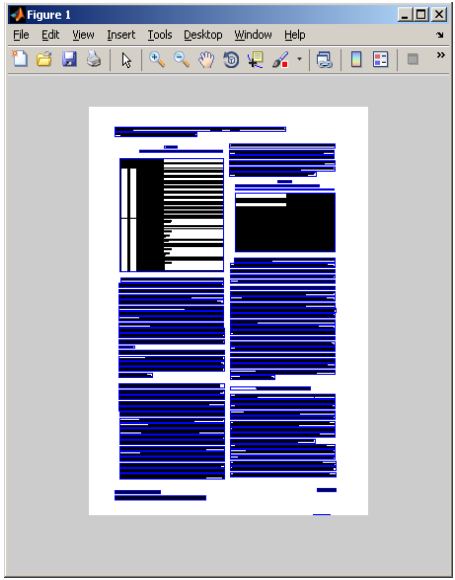

Fig. 3. Labeling Connected Components

### D. Feature Extraction

Feature extraction is the process of transforming the document image block into set of features. The features describing the properties of each block are extracted based on its edges. The coordinates of the edges, specify the size of the block. Each extreme edges of block are defined by considering the origin of binary image.

Features such as height, area, aspect ratio, perimeter, perimeter/height ratio, average horizontal length are extracted to differentiate the non-text block from text block. Image, drawings and table blocks are identified by means of following features: density, density of segmented block, horizontal transition along x-axis and y-axis, vertical transition along x-axis and y-axis. The table and drawings blocks can be recognized more accurately using the features such as mean standard deviation, active pixels, energy and entropy.

Blocks are represented by (Xmin, Ymin) and (Xmax, Ymax).Where,

Ymin - Left-most pixel value of column.

Ymax - Right-most pixel value of column.

Xmin - Left-most pixel value of row.

Xmax - Right-most pixel value of row.

The following block features are computed for classification of document region.

Height (H) - Height of the block is computed by subtracting the leftmost pixel from rightmost pixel of column.

$$H = D_x = (Y_{max} - Y_{min}) + 1$$

Width (W) - Width of the block is computed by subtracting the leftmost pixel from rightmost pixel of row.

$$W = D_y = (X_{max} - X_{min}) + 1$$

Aspect Ratio (E) - It is defined as the ratio of a block's width-to-height.

$$E = W/H$$

Area (A) - Area of the block is obtained by multiplying the height and width.

$$A = H*W$$

Density (D) - Density is defined as the ratio of total number of black pixels within each block of document image to the area.

$$D = N/Area$$

Horizontal transition along x axis (HTx) - It is defined as ratio of horizontal transitions per unit height and computed as

$$HT_x = HT/D_x$$

Where, HT is the horizontal transitions of black to white or white to black pixels in a block of document image.

Vertical transition along x axis (VTx) - It is defined as ratio of vertical transition per unit height and is computed using the following formula.

$$VT_x = VT/D_x$$

Where, VT is the vertical transitions of black to white or white to black pixels in a block of document images.

Horizontal transition along y axis (HTy) - It is defined as horizontal transition of white to black pixels per unit width and is given by

$$HT_y = HT/D_y$$

Vertical transition along y axis (VTy) - It is defined as vertical transition of white to black pixels per unit width and is given by

$$VT_y = VT/D_y$$

Density of segmented block (D1) - It is the ratio of total number of black pixels of segmented block after applying Run length smearing algorithm to the area.

$$D1 = C/Area$$

Average Horizontal Length(R) - It is defined as mean horizontal length of black runs of document image within each block.

$$R = N/HT$$

The Product of block height and ratio R is given by

$$RH = R*H$$

The Product of ratio R and ratio E is given by

$$RE = R*E$$

The Product of ratio D1 and ratio R is given by

$$RD = R*D1$$

Mean (E) - Mean value of pixel intensities in the block is defined as arithmetic average of distribution of pixels in the block. It is computed using the function mean () in mat lab.

$$E_i^{n-1} = \sum x_i / N$$

Standard Deviation (σ) - Standard deviation of pixel intensities in the block is defined as calculated using the function std () in mat lab.

$$\sigma_i^{n-1} = \sum (x - \mu)$$

Active pixels (A) - Active pixels of the block is defined as the count of pixels with intensity < mean-k*std and it is computed as follows.

$$A = Sum\ (I < (mean-k*std))$$

Perimeter (P) - Perimeter is defined as the distance around the block and computed using the formula

$$P = 2*(H+W)$$

Ratio S - It is defined as ratio of perimeter to height of the block and is given by

$$S = Perimeter/height$$

Energy (E) - It is defined as sum of squared elements in the image and is also known as uniformity or the angular second moment.

$$E_{i=1}^n = \sum x_i^2$$

Entropy- It is defined as statistical measure of randomness that can be used to characterize the texture of the input image. It is calculated using the function entropy () in mat lab.

$$\text{Entropy} = \sum_{i,j} P(i,j) \log P(i,j)$$

Thus a group of 20 distinct features are extracted from the blocks of document image using mat lab. The feature vectors are generated using 519 segmented blocks of 15 document pages and the training dataset is generated with 519 instances.

### III. LINEAR GENETIC PROGRAMMING-GENETIC PROGRAMMING MODEL

In artificial intelligence, genetic programming (GP) is an evolutionary algorithm based methodology impressed by biological evolution to seek out computer programs that carry out a user-defined task. It is an area of genetic algorithms (GA) where each entity is a computer code. It is a machine learning technique used to optimize a population of computer programs according to a fitness landscape determined by a program's ability to perform a given computational task[14].

#### A. GP algorithm

GP algorithm involves few essential control parameters for solving the problem. The control parameters needed are population size, maximum number of generations, solution fitness and operators. The algorithm begins by initializing the desired population size and evaluates fitness until best solution found. Crossover, reproduction and mutation are the common operators used in GP.

#### B. GP operators

Crossover operator works by choosing two parents from the population. Two random sub trees are selected from every parent and swapped to make children. In mutation single node in parent tree is chosen and replaced with a random node of same sort. E.g. a function node is replaced by a function node of same sort and a terminal node is replaced by a randomly selected terminal node. In reproduction operator an individual is chosen and copied on to the new generation with none changes or modifications to that.

#### C. Classification Using Genetic Programming

The GP based classification methods are categorized into three types. First kind of method illustrates the evolution of classification algorithms like decision trees, neural networks or other rule based algorithms. This system accurately depicts the use of GP for program evolution. The second system contains evolution of classification rules or expressions. Rules are evolved in the type of logical expressions with logical operators. In the third system the expressions are evolved in the type of arithmetic expressions or functions [15].

#### D. Fitness function for classification

One among the most common fitness function for classification task is the classification accuracy. The accuracy tells the amount of instances properly classified by a classifier. Another measure is to minimize the classification error that is reciprocal of classification accuracy. Fitness measure for GP classification task is uncomplicated. It is determined by the classification accuracy that is expressed as the following formula.

$$fitness = \frac{Correctly\ classified\ instances}{Total\ no\ of\ instances} \times 100$$

#### E. Linear Genetic Programming

Linear genetic programming is one type of representation of GP. In Linear genetic programming LGP each program is a sequence of register instructions and usually expressed in human-readable form. The registers are assigned zero before program execution. The features representing the objects to be classified are loaded into registers. The program is executed in an imperative manner and represents a directed acyclic graph. Fig 4 shows the program generated by linear genetic programming based discipulus. Discipulus has a feature called self tuning and self parameterizing which selects the GP control parameters based on its problem to be solved.

```
if (cflag) f[0] = f[0];
f[0] = - f[0];
tmp = f[0]; f[0] = f[0]; f[0] = tmp;
f[0] * = Input8;
tmp = f[1]; f[1] = f[0]; f[0] = tmp;
cflag = (Double.isNaN(f[0]) || Double.isNaN(f[0])) ? true :
(f[0] < f[0]);
tmp = f[0]; f[0] = f[0]; f[0] = tmp;
 f[0]* = Math.pow(2,trunc(f[1]));
cflag = (Double.isNaN(f[0]) || Double.isNaN(f[1])) ? true :
(f[0] < f[1]);
f[0] = Math.cos(f[0]);
f[0] = Math.sin(f[0]);
f[0] = Math.abs(f[0]);
f[0]+ = 1.366016626358032f;
f[0]* = -1.924433708190918f;
f[0] = Math.abs(f[0]);
f[0]+ = f[0];
f[0]* = -1.924433708190918f;
f[0]- = Input17;
f[1]- = f[0];
cflag = (Double.isNaN(f[0]) || Double.isNaN(f[0])) ? true :
(f[0] < f[0]);
f[0]+ = Input1;
f[0]- = -0.494312047958374f;
 f[0]* = -1.924433708190918f;
f[0]/ = Input17;
f[0]/ = -0.494312047958374f;
 f[0] = Math.abs(f[0]);
f[1]* = f[0];
f[0]/ = -0.494312047958374f;
 f[0]/ = Input3;
f[0]/ = Input1;
f[0]/ = Input2;
f[0]* = 1.252994060516357f;
f[0]- = -0.494312047958374f;
cflag = (Double.isNaN(f[0]) || Double.isNaN(f[0])) ? true :
(f[0] < f[0]);
tmp = f[0]; f[0] = f[0]; f[0] = tmp;
cflag = (Double.isNaN(f[0]) || Double.isNaN(f[0])) ? true :
(f[0] < f[0]);
tmp = f[0]; f[0] = f[0]; f[0] = tmp;
cflag = (Double.isNaN(f[0]) || Double.isNaN(f[0])) ? true :
(f[0] < f[0]);
f[0]* = f[0];
return (float) f[0];
}
```

Fig. 1.    Program Generated By LGP Using Discipulus Tool

Where input[i] represents the input variables, f[i] represents binary class values used in LGP. The function set of the system consists of arithmetic operators (=,,-,/,*), conditional operator (if condition) and trigonometric measures (sin, cos, tan, log, sqrt etc). The output of the program is decided by considering the highest value hold by f[i] register.

IV. EXPERIMENT AND RESULTS

The document image classification model is generated by implementing supervised learning algorithm. The documents used for creating the dataset are collected from various journals and here 15 document pages have been used. The dataset consists of 519 segmented blocks of which 196 are text blocks, 123 are drawing blocks, 92 are image blocks and 108 are tabular blocks. The features describing the properties of the blocks are extracted and the size of each feature vector is 20. The dataset with 519 instances is trained by genetic programming using discipulus tool.

Discipulus is the world's first and fastest commercial Genetic programming and data analysis software. Discipulus writes computer programs automatically in Java, C, C Sharp, Delphi and Intel assembler code, all on a desktop computer. Discipulus handles only binary classification. But, the problem of document segmentation is a multiclass classification. Hence, it has been solved by extending binary classification into multiclass classification using one against one method.

A. One Against One

In this case, for 4 classes (text, image, drawings and table) 6 multiclassifiers have been built, one for each pair of classes. The prediction of class for a new data point is based on voting scheme. Fig 5 shows the architecture of the prediction scheme for the four classes.

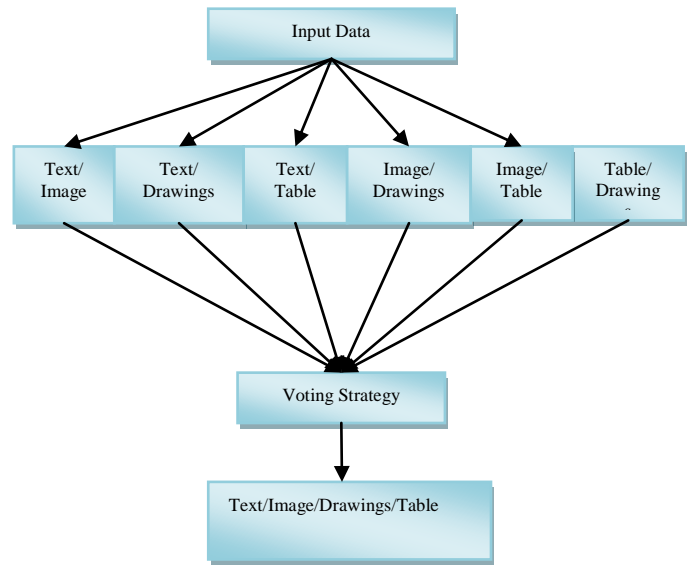

Fig. 5.    Prediction Using One Against One Multiclassification

B. Training and testing in Discipulus

Discipulus is bundled with Notitia, which is data preparation, cleaning, and import software. Notitia lets you import, clean-up, transform and split data for use in Discipulus. Fig 6 and Fig 7 shows data import, data preparation using Notitia and data export to Discipulus respectively. For each binary model, hits then fitness approach is used as fitness function for classification.

In hits then fitness approach, different weights (or costs) are assigned to positive and negative hits to obtain hit rate. Positive hit is the hit rate assigned to positive instance (1). Negative hit rate is the hit rate assigned to negative instances (0).

The hit-rate is defined as the percentage of correctly classified instances. Threshold value is used to classify the outputs of binary models. If program's output for an instance is greater than or equal to the classification threshold, that output is counted as class one output. Else if the program's output for an instance is less than the classification threshold, that output is counted as class zero output. Threshold value is set automatically halfway between the target outputs for classes zero and one.

The dataset is used for learning six binary classifiers are built as per one against one multiclassification scheme. The java codes corresponding to these six classifiers or classification models have been saved using Discipulus. These six models have been invoked in a java program and voting strategy is implemented to test the class label of the given input vector. In this manner a test dataset with 40 instances, 10 in each class (text, image, drawings, and table) is tested and classification accuracy is calculated. It is observed that about 97.5% of classification accuracy is shown by genetic programming based document segmentation and region classification. Fig 8 to 13 depict the training and testing of GP based multiclassification in Discipulus

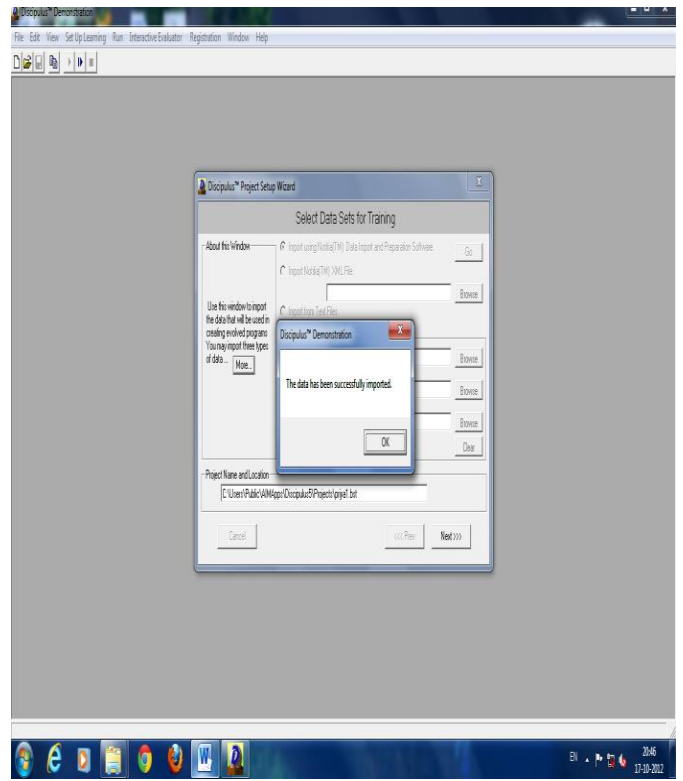

Fig. 7.     Data Export Using Notitia To Discipulus

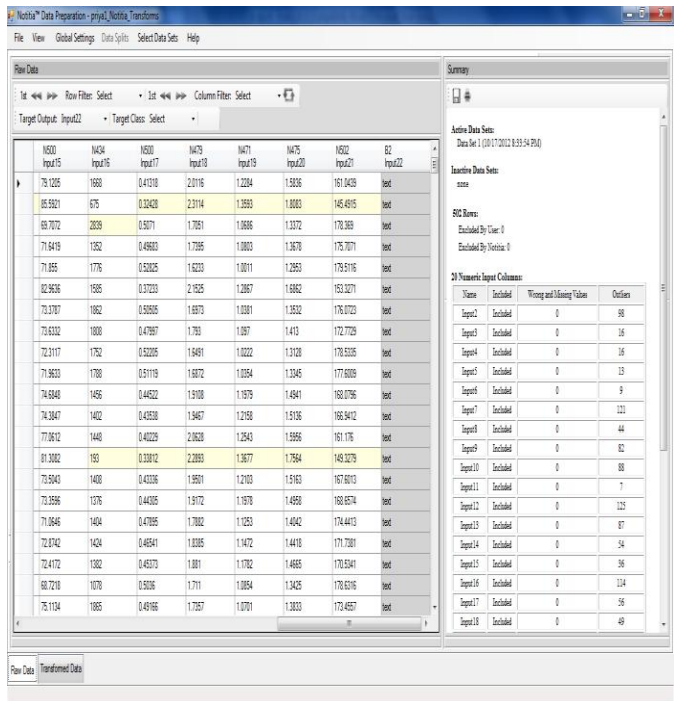

Fig. 6.     Data Import And Preparation Using Notitia

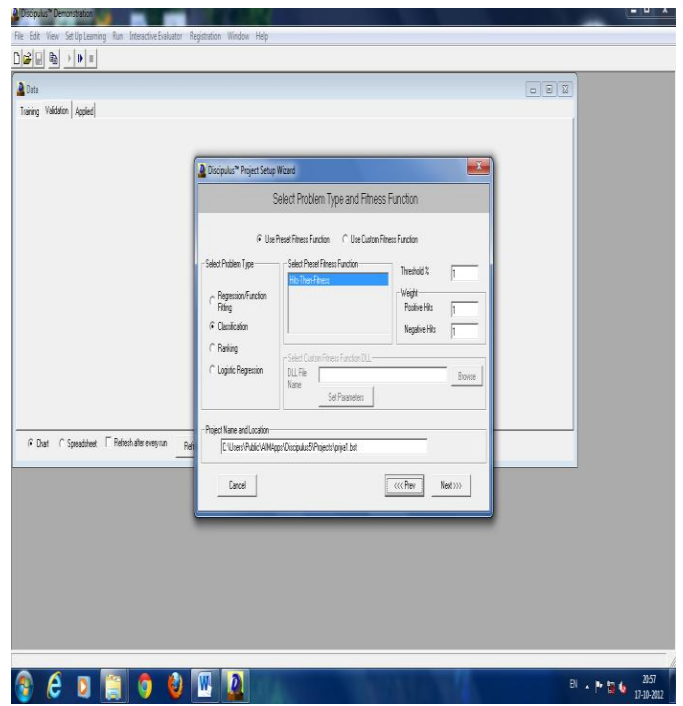

Fig. 8.     Classification On Discipulus

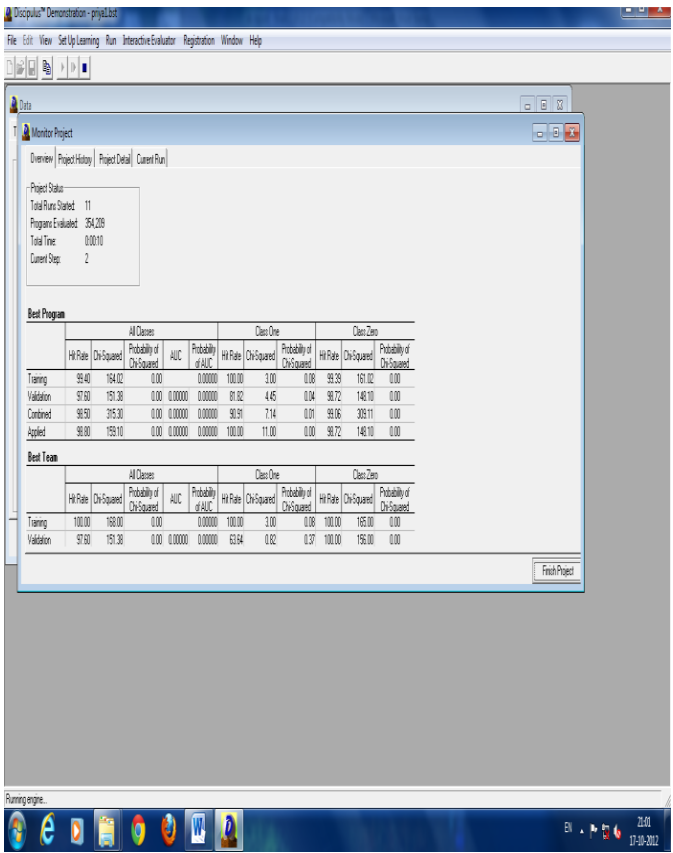

Fig. 9. Project Status On Discipulus

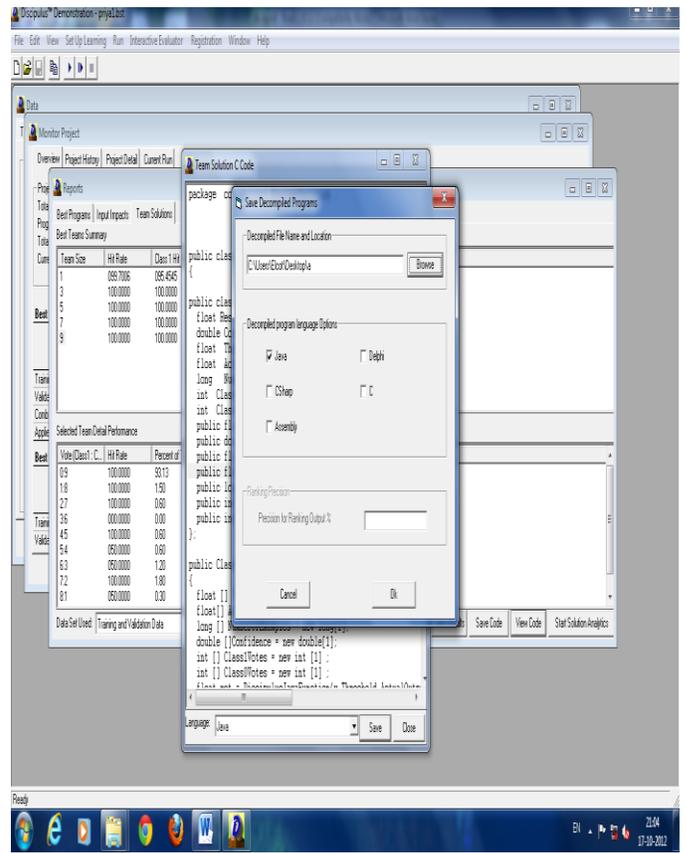

Fig. 11. Saving Of Java Code

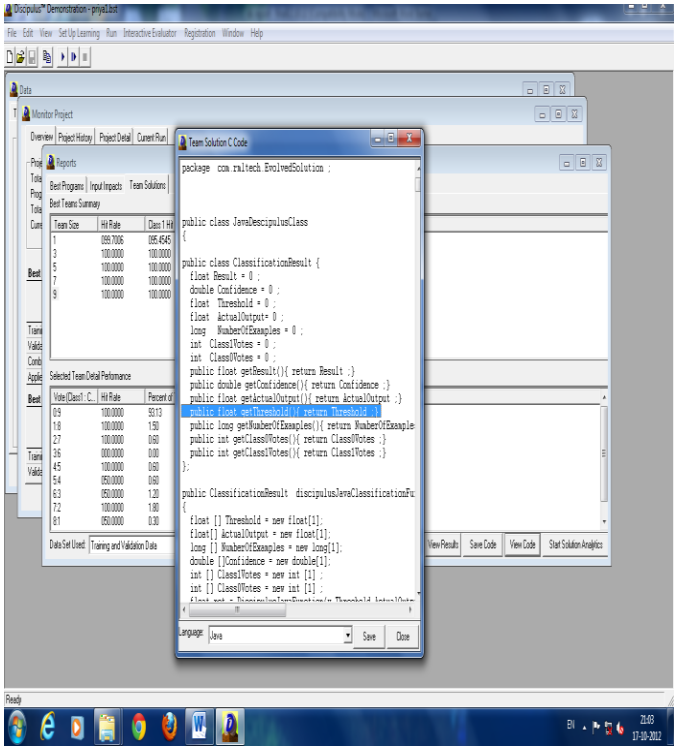

Fig. 10. Java code for classification model created using Discipulus

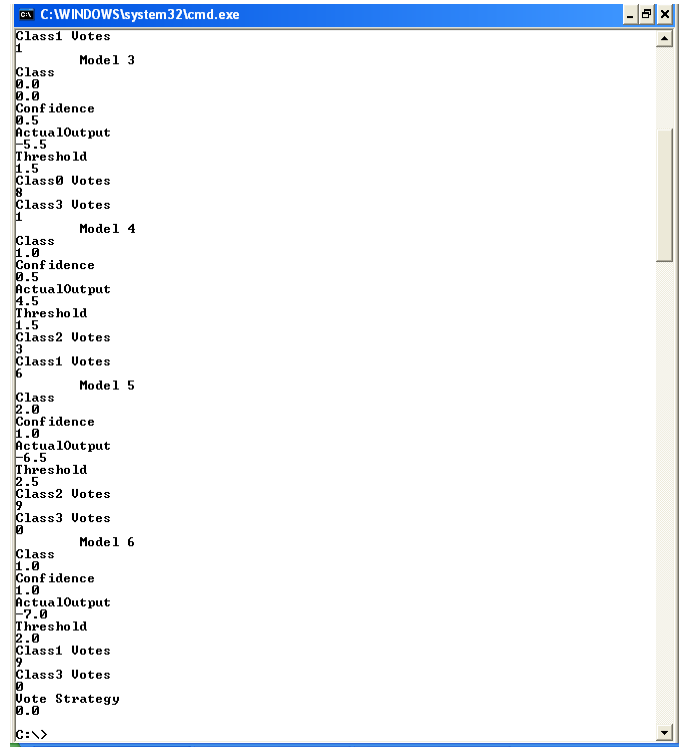

Fig. 12. Result Of Testing New Dataset In Java

Fig. 13. Cross Validation Of GP

## V. CONCLUSION

This paper demonstrates the modeling of document segmentation as classification task and describes the implementation of genetic programming approach for classifying various regions. Discipulus have been applied for generating LGP based classification models. It is observed that classification implemented by genetic programming in this paper is more efficient than other machine learning algorithms because the commercial GP software Discipulus uses automatic induction of binary machine code to achieve better performance.

Document segmentation model constructed in this research work can be incorporated into "Form" processing system to extract the data which are presented in text, image, table, drawing format from any kind of "Forms". Examples of "Forms" may include a student application form, bank application form.

AUTHORS PROFILE

**N. Priya dharshini** is pursing Master of Philosophy in Computer Science in PSGR Krishnammal college for women under the guidance of MS.Vijaya. Her research interests are data mining, image processing, and pattern recognition.

**MS. Vijaya** is presently working as Associate Professor in GR Govindarajulu School Of Applied Computer Technology, PSGR Krishnammal college for women, Coimbatore, India. She has 22 years of teaching experience and 8 years of research experience. She has completed her doctoral programme in the area of Natural Language Processing. Her areas of interest include Data Mining, Support Vector Machine, Machine learning, Pattern Recognition, Natural Language Processing and Optimization Techniques. She has presented 22 papers in National conferences and she has to her credit 17 publications in International conference proceedings and Journals.

She is a member of Computer Society of India, International Association of Engineers (Hong Kong), International Association of Computer Science and Information Technology (IACSIT – Singapore). She is also a reviewer of International Journal of Computer Science and Information Security.